\documentclass[11pt,letterpaper]{article}
\usepackage{emnlp2017}
\usepackage{times}
\usepackage{latexsym}

\usepackage{url}
\usepackage{mathtools}
\usepackage{amsmath}
\usepackage{amssymb}
\usepackage{bm}
\usepackage[ruled,linesnumbered]{algorithm2e}
\usepackage{multirow}
\usepackage[font={normalsize}]{caption}
\usepackage{subcaption}
\usepackage{graphicx}
\usepackage{esvect}
\usepackage{adjustbox}
\usepackage{hypcap}

\usepackage{tikz}
\usetikzlibrary{arrows}
\usetikzlibrary{calc}
\usetikzlibrary{shapes.geometric}

\usetikzlibrary{positioning}
\usepackage{tkz-graph}

\usepackage{float}
\floatstyle{plaintop}
\restylefloat{table}

\usepackage[belowskip=-15pt,aboveskip=0pt]{caption}

\emnlpfinalcopy

\def\emnlppaperid{1420}

\title{Text Classification with Novelty Detection}

\author{Qi Qin$^{1,2}$ , Wenpeng Hu$^{3}$, Bing Liu$^{4,}$\thanks{\ \ Corresponding Author.}\\
$^1$ Center for Data Science, AAIS, Peking University\\
$^2$ Wangxuan Institute of Computer Technology, Peking University\\
$^3$ Department of Information Science, Peking University\\
$^4$ Department of Computer Science, University of Illinois at Chicago\\
\{qinqi, wenpeng.hu\}@pku.edu.cn, liub@uic.edu
}

\date{}

\begin{document}
	\maketitle
	\begin{abstract}
		
		This paper studies the problem of detecting novel or unexpected instances in text classification. In traditional text classification, the classes appeared in testing must have been seen in training. However, in many applications, this is not the case because in testing, we may see unexpected instances that are not from any of the training classes. In this paper, we propose a significantly more effective approach that converts the original problem to a pair-wise matching problem and then outputs how probable two instances belong to the same class. Under this approach, we present two models.~The more effective model uses two embedding matrices of a pair of instances as two channels of a CNN. The output probabilities from such pairs are used to judge whether a test instance is from a seen class or is novel/unexpected. Experimental results show that the proposed method substantially outperforms the state-of-the-art baselines.
		
	\end{abstract}
	
\section{Introduction}
	
Traditional text classification \citep{Kim2014Convolutional,tang2015document,joulin2017bag,shen2018baseline,2020Feature} assumes a closed world where all classes appear in the test data must have appeared in the training data~\citep{fei2016breaking}. However, this assumption is not true in many real-world applications. For example,an intelligent personal assistant (e.g., Amazon Alexa, Microsoft Cortana) 
needs to classify user utterances into existing known intent classes and also detects or rejects utterances with unknown intents. To effectively work in such dynamic and unpredictable environments, the learned model has to be able to classify instances belong to the old/seen classes and also spot novel or unexpected instances of some new/unseen classes. This problem is called \textit{open-world classification} (\textbf{OWC}) ~\citep{bendale2015towards,fei2016breaking}.
	
\textbf{Problem Definition:} Given a training set ${D} = \{(x_i, y_i)\}_{i=1}^{N}$, where $x_i$ is a train example or instance, $y_i\in C_s = \{c_1,...,c_m\}$ is $x_i$'s class label and $m$ is the total number of seen/known classes in $D$. The test data $T$ is from the classes in $C_s \cup C_u$, where $C_u$ is a set of hidden unseen classes. Our goal is to build a classifier $F$ from $D$ and test it on $T$ so that $F$ can classify each instance in $T$ to its correct class in $C_s$ or reject it as belonging to a novel unseen class in $C_u$. Since we don't know the classes in $C_u$, we use $c_{rej}$ to represent them all. Previous approaches typically used each instance as an input to train a classifier, which needs a large number of network parameters to remember the characteristics of the training/seen classes. In this paper, we propose a much more effective method based on pair-wise matching. We call it Pairwise Matching Network (\textbf{PM-Net}). Specifically, it learns to estimate the probability that two given instances belong to the same class. For training, we first use the original training data $D$ to create a new pair-wise training dataset $D^{'}= \{(x_{1k},x_{2k},y_k)\}_{k=1}^{M}$, where $x_{1k}, x_{2k} \in D$ with $y_k$ being the label. If $x_{1k}$ and $x_{2k}$ are from the same class in the original data $D$, $y_k = 1$; otherwise $y_k = 0$. We then use $D^{'}$ to train the matching function $f(x_{1k}, x_{2k})$ to estimate the probability $p$ of $x_{1k}$ and $x_{2k}$ belonging to the same class. In testing, for each test instance $x_t \in T$, we first construct a memory $M_i$ for each seen class by randomly selecting $K$ examples from the class $c_i$ in $D$ and put them in $M_i$. Then, $x_t$ forms a pair with every example in the whole memory $M = \{M_i\}_{i=1}^m$ to calculate the probability that $x_t$ has the same class as the compared example using the matching function $f(\cdot ,\cdot)$. Next, we calculate the mean probability
of the $K$ probabilities of each seen class $c_i$ after the maximum and minimum values are removed to obtain $x_t$'s average probability $p_i$ of belonging to class $c_i$ (we also experimented with a few other strategies, but they were poorer). After we get the probability of $t$ belonging to every seen class, we have $P = \{p_{1}, p_{2},...,p_{m}\}$. Finally, we use Eq.~1 to determine whether $t$ belongs to one of seen classes or is a novel instance from the unseen class $c_{rej}$~\footnote{Like DOC, we use Gaussian fitting to estimate the threshold.}. 
\begin{equation}
  \label{eq:alpha}
   \hat{y} =\left\{           
      \begin{array}{lcl}    
         c_{rej},  &if  \mathop{\max} \limits_{c_i} {P} < threshold; \\   
         \mathop{\arg \max} \limits_{c_i} {P}, &otherwise.
      \end{array}
    \right.	
\end{equation}

We propose two models (Figure 1) to learn the matching function: 
1) PM-Net 1 extracts an advanced feature vector from each input instance in a pair individually using a convolutional neural network (CNN), and then concatenates the two feature vectors to learn a matching score (\emph{i.e.}, probability) using the matching score module; 
2) PM-Net 2 uses the embedding matrices of the two input instances as the two channels of the CNN extractor. Then the matching score module outputs a probability. We can see that PM-Net 1 can only extract instance-level interactions between the two instances, but PM-Net 2 can obtain more fine-grained interaction information (\emph{e.g.}, word-level, phrase-level and sentence-level). Extensive experiments show that the proposed approach (both models) outperforms strong baselines considerably. Comparing with PM-Net 1, PM-Net 2 is more effective.

\section{Related work}
Open-world classification has been studied in text classification and computer vision (where it is called open-set recognition). In text classification, one-class SVM \citep{scholkopf2001estimating} is the earliest method, which performed poorly because it didn’t use negative examples. Also, it doesn't do multi-class classification.~\citet{fei2016breaking} proposed a Center-Based Similarity space learning method to reject/accept a test instance by deciding whether it is outside the decision hypersphere of each class. \citet{fei2016learning} further added the capability of incrementally learning the new classes. \citet{shu2017doc} outputted the probability that a test instance belongs to each of the seen classes by an 1-vs-rest layer. It then compares the probability value with an estimated threshold to determine whether a test instance is unexpected or not. \citet{xu2019open} combined $k$NN and meta-learning to solve the problem. \citet{Zheng2019Out} used autoencoder and adversarial training to detect out-of-domain sentences in dialogue systems. 

Although our work focuses on text classification, related works have also been done in computer vision. For example, \citet{scheirer2012toward} recognizes unseen images by reducing open space risk. \citet{jain2014multi} performed similar tasks.
\citet{bendale2016towards} introduced an OpenMax layer to adapt a deep network for OWC. It has been shown that these methods are poorer than the DOC method in \citep{shu2017doc,xu2019open} for text. However, to the best of our knowledge, none of these known methods consider a pair-wise model with a reference class instance to build an OWC classifier.

\begin{figure}[t]
  \centering
  \includegraphics[width=1.0\columnwidth]{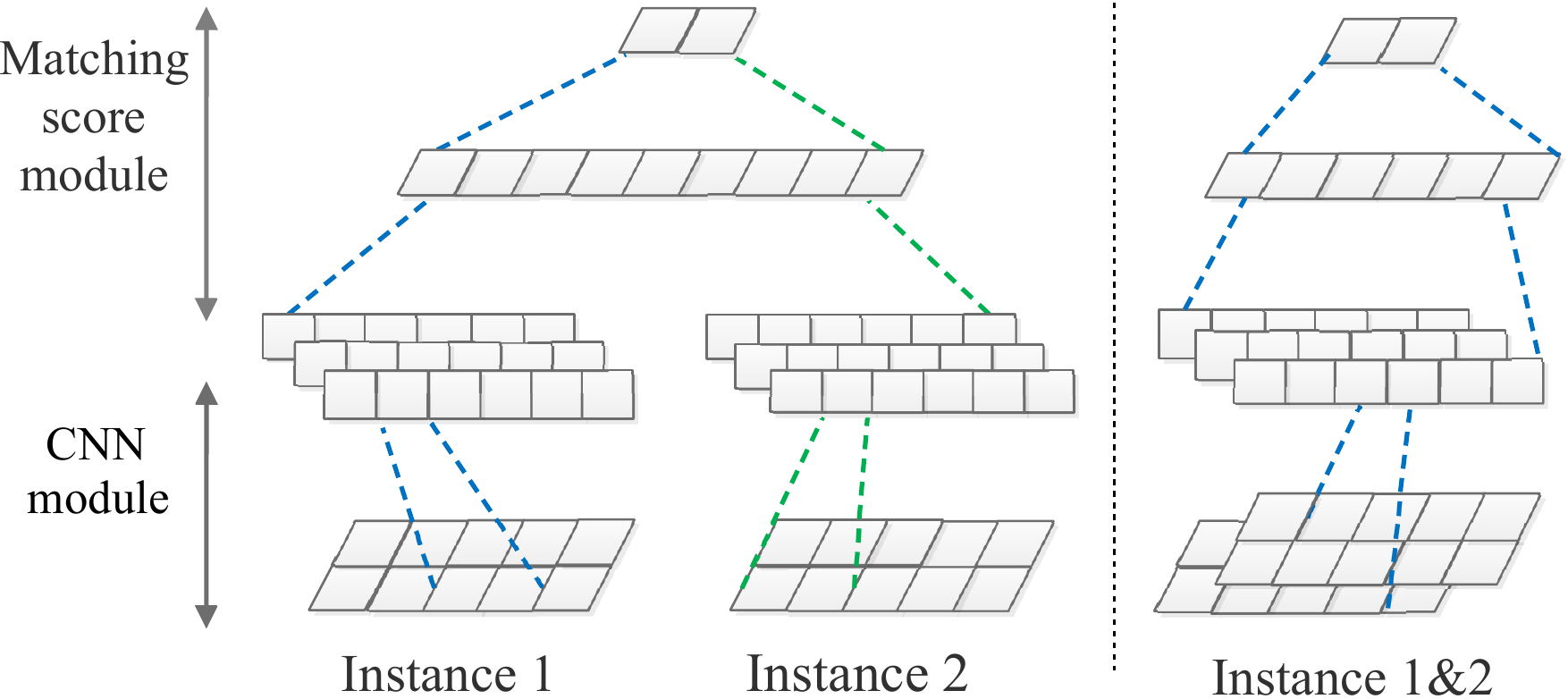}
  \caption{Two models of the matching function: PM-Net 1 (left) and PM-Net 2 (right).}
\end{figure} 
	
\section{Models}
The proposed PM-Net (\emph{i.e.}, Figure 1) uses the CNN architecture \citep{Kim2014Convolutional}. Next, we briefly describe the CNN model. Given an instance $x$ (after padding and cutting) with length $L$, let $X \in \mathbf{R}^{L\times H}$ be the embedding matrix of $x$, where $H$ is the embedding size. 
To get more fine-grained features, we use different filter sizes $[n,H]$ and concatenate the feature maps of different filter sizes as the final representation. We refer to the CNN module as
\begin{equation}
       r = \mbox{CNN}(X)
\end{equation}
PM-Net 1 and PM-Net 2 are presented next. 
	
\subsection{PM-Net 1}
As mentioned earlier, we need a matching function $f(x_{1k},x_{2k})$ to compute the probability $p$ of two instances $x_{1k},x_{2k}$ belonging to the same class. In PM-Net 1, we first use the CNN module to obtain $x_{1k}$ and $x_{2k}$’s advanced representation vectors, which are
\begin{equation}
    r_{1k} = \mbox{CNN}(X_{1k}),r_{2k} = \mbox{CNN}(X_{2k}).
\end{equation}
where $X_{1k}, X_{2k} \in \mathbf{R}^{L\times H}$ is $x_{1k}$, $x_{2k}$'s embedding matrix and $r_{1k}$, $r_{2k} \in \mathbf{R}^d$. Then, we concatenate $r_{1k}$ and $r_{2k}$ as the interaction features of $x_{1k}$, $x_{2k}$, which are fed to two fully connected layers,
\begin{equation}
    fc = \mbox{Relu}( r_{1k} \oplus r_{2k} \cdot W_1 + b_1)
\end{equation}
\begin{equation}
    \hat{y_k} = \mbox{Softmax}( fc \cdot W_2 + b_2)
\end{equation}
where $\oplus$ is the concatenation operation, $W_1
,W_2 
$ are the weights of fully connected layers and $b_1 
,b_2 
$ are bias. Note that $\hat{y_k}$'s second element indicates the probability that $x_{1k}$ and $x_{2k}$ belong to the same class. The loss function of PM-Net 1 is computed based on the actual label $y_k$ and $\hat{y_k}$, $k = 1,...,M$. Let $Y = \{y_k\}_{k=1}^M$ and $\hat{Y} = \{\hat{y_k}\}_{k=1}^M$. PM-Net 1's loss function is 
\begin{equation}
    Loss_1 = \mbox{CrossEntropy}(Y,\hat{Y})
\end{equation}

\subsection{PM-Net 2}
From Eq.~3, we can see that PM-Net 1 only covers the interaction of the two instances at the instance level, which may not capture the fine-grained information. PM-Net 2 wants to capture dependency information of the two input instances at more fine-grained levels (\emph{i.e.}, word-level, phrase-level and sentence-level), which allow the matching function to output more accurate probabilities. In order to achieve this goal, we propose a novel matching model, which combines the two input instances $x_{1k}$, $x_{2k}$'s embedding matrices $X_{1k}$, $X_{2k} \in \mathbf{R}^{L\times H}$ into a three-dimensional matrix $X = X_{1k} \oplus X_{2k}$, $X \in \mathbf{R}^{L\times H \times 2}$,
from the input layer. Then, through the CNN module we can obtain $x_{1k}$, $x_{2k}$’s multi-granular interactions or dependencies. We then have
\begin{equation}
    r_k = \mbox{CNN}(X)
\end{equation}
Like PM-Net 1, we obtain the probability of the two instances belonging to the same class after two fully connected layers.
\begin{equation}
    fc = \mbox{Relu}(r_k \cdot W_3 + b_3)
\end{equation}
\begin{equation}
    \hat{y_k} = \mbox{Softmax}(fc \cdot W_4 + b_4)
\end{equation}
where $W_3 
,W_4 
$ and $b_3 
,b_4 
$ are parameters. The loss function of PM-Net 2 and PM-Net 1 are the same.
By using PM-Net, a test instance will get the probability of belonging to each seen class. Then we use these  probabilities to compare with an estimated threshold (\emph{i.g.}, Eq.~1) to judge whether the test instance is from a seen class or an unseen class.
\section{Experiments}
\subsection{Datasets}
To evaluate the effectiveness of the proposed models, we conducted experiments using two datasets:

\textbf{THUCNews}: THUCNews contains 14 classes and 836,062 news articles \citep{sun2016thuctc}.\footnote{\small\tt http://thuctc.thunlp.org} 

\textbf{News Category Dataset (NCD)}: This dataset contains around 200k news headlines from the year 2012 to 2018 obtained from HuffPost.\footnote{\small\tt https://rishabhmisra.github.io/
publications/}


Note that we only use 10 classes in each dataset. This is because the baseline L2AC needs additional classes for meta-learning. 
Both datasets contain only the training and test sets and the test set is randomly drawn 200 instances from each class. THUCNews's training set $D$ contains 15,000 instances per class. NCD's training set $D$ contains 3,000 instances per class.

\begin{table*}[t!]
\begin{center}
\scalebox{0.8}{
\begin{tabular}{l||c|c|c|c|c|c}
\hline
\multirow{3}*{}  &
\multicolumn{3}{c|}{THUCNews}&\multicolumn{3}{c}{NCD}\\
\cline{2-4}
\cline{5-7} 
& 3:7 & 5:5 & 7:3 & 3:7 & 5:5 & 7:3\\ \hline
DOC-LSTM & 66.58 &64.56 &69.29 &51.25 & 53.06 &58.46\\ 
DOC-CNN & 66.62 & 66.98 & 69.20 & 54.04 & 53.13 & 58.62\\ \hline
L2AC & 65.66 & 67.47 & 74.59 & 40.95 & 48.35 & 49.08\\ \hline
\hline
PM-Net 1-K=1 &74.77 &81.76 &74.60	&61.35	&66.29 &62.73 \\ 
PM-Net 1-K=5 &77.76 &83.85 &80.02 &61.19	&69.16 &67.63 \\  
PM-Net 1-K=15 &79.21 &84.43 &79.62	&62.26 &70.49 &69.47 \\ 
PM-Net 1-K=100 &78.67	&83.37 &76.64 &61.49 &71.56 &68.74\\
\hline 
PM-Net 2-K=1 &77.05 &80.84 &78.09 &61.56	&66.80 &64.25 \\ 
PM-Net 2-K=5 &79.26 &83.26 & 80.86 &62.57 &71.41 &68.93 \\
PM-Net 2-K=15 &81.81 &83.06 &81.18 &63.24 &72.22 &70.82 \\ 
PM-Net 2-K=100 &80.01	&84.18 &81.13 &65.36 &73.19 &71.82\\
\hline 
\end{tabular}
}
\end{center}
\caption{All results in Macro-F1(\%). 3:7, 5:5 or 7:3 is the ratio of the seen classes to unseen classes. $K$=1, 5, 15, or 100 is the size of each memory dataset $M_i$ for each seen class. 
}
\label{tab:accents}
\end{table*}

\subsection{Baselines} 
We use DOC \citep{shu2017doc} and L2AC \citep{xu2019open} as our baselines.
To the best of our knowledge, they are the state-of-the-art systems for open-world text classification. It has been shown in~\citep{shu2017doc} that DOC significantly outperforms the other methods CL-cbsSVM and cbsSVMin  \citep{fei2016breaking} and OpenMax \citep{bendale2016towards}. We will not compare with them. 

\textbf{DOC-CNN}: This is the original DOC with Gaussian fitting to set the threshold for rejection.

\textbf{DOC-LSTM}: This is a variant of DOC-CNN. We use LSTM to replace CNN to encode the input sequence. The hidden state size of LSTM is 512.

\textbf{L2AC}: We use the hyper-parameters: $k$ = 5 for $K$NN (which gives the best results), and $n$ = 9 for the meta-classifier’s negative classes. For THUCNews, since there are only 4 classes left for L2AC's meta-learning, we use $n$ = 3 (\emph{i.e.}, each positive sample with three negative samples) for this dataset.

\subsection{Implementation Details}
In our experiments, pre-trained embeddings were used for all models, including the baselines. In the experiments related to THUCNews and NCD, we used separately a 200-dimensional vector representation released by Tencent \footnote{\small\tt https://ai.tencent.com/ailab/nlp/embe-
dding.html} and Google's pre-trained 300-dimension word embeddings \footnote{\small\tt https://code.google.com/p/word2vec/} respectively.\footnote{Note that many other embeddings can be used in our system\cite{pennington2014glove,hu2016different} } All models' CNN modules use filters with window size $n$ in $[3,4,5]$ and each filter window with 100 feature maps. 

In the evaluation, we hold out some classes as unseen in training and mix them back during testing. We vary the number of seen classes 3, 5 and 7 (total number of classes is 10) for training and all 10 classes are used in testing. We use 7:3 as an example to detail the data preparation. First, we randomly select 7 classes as the seen classes and the rest 3 classes as the unseen/novel classes. Then, we build a new pair-wise training dataset (\emph{i.e.}, $D^{'}$). For each instance $x_i$ in the training data $D$ of the seen classes, we randomly select another instance $x_j$ from the class of $x_i$ to produce a positive example ($x_i$, $x_j$, 1). In the remaining classes (6 of them), we first randomly sample a class and then an instance $x_k$ from the class to produce a negative example ($x_i$, $x_k$, 0). In testing, we select $K$ examples from each of the 7 seen classes as the memory $M_i$. For each test instance $x_t \in T$ (test set), we use PM-Net to produce the average probability that $x_t$ belongs to each of the seen classes. Finally, we use Eq.~1 to classify $x_t$ to one of the seen classes or reject it as a novel instance. 

\subsection{Experiment Results}
The results of THUCNews and NCD are given in Table 1. We use macro F1-score for evaluation. From Table 1, we can observe the following:

First, our two models perform considerably better than DOC and L2AC in macro-F1 scores for both datasets in the 3:7, 5:7, and 7:3 cases. Even when the size $K$ of the memory for each seen class is 1, our models still perform much better than the baselines. For example, in the 3:7 case on THUCNews, our PM-Net 1 and PM-Net 2 are $9.11$ and $11.39$ higher than L2AC respectively.

Second, as the memory size $K$ increases, we get higher macro-F1, which is expected because $K=1$ can be quite unreliable. When $K>1$, we remove the maximum and minimum values of each test instance belonging to each seen class, which can eliminate the effect of singular values to give us a more reliable information of the seen classes, and help us decide if a test instance should belong to a seen class or be rejected as novel. 

Third, comparing the last two blocks in Table 1, we can see that using the two embedding matrices as the CNN's input channel in PM-Net 2 is better than PM-Net 1. This shows that combining two instances in the first layer makes it easier to extract more fine-grained features for classification.

\section{Conclusion}
This paper proposed a novel method to solve the OWC problem for text, which converts traditional OWC to a pair-wise matching problem. 
Using THUCNews and NCD datasets, the paper showed that the proposed two models perform dramatically better than the state-of-the-art baselines. In our future work, we plan to further improve the accuracy. 

\bibliography{emnlp2017}
\bibliographystyle{emnlp_natbib}

\end{document}